\documentclass[11pt,a4paper]{article}

\usepackage[utf8]{inputenc}
\usepackage[T1]{fontenc}
\usepackage{lmodern}
\usepackage{microtype}
\usepackage[margin=1.0in]{geometry}
\usepackage{amsmath,amssymb,amsthm}
\usepackage{mathtools}
\usepackage{booktabs}
\usepackage{tabularx}
\usepackage{multirow}
\usepackage{graphicx}
\usepackage{xcolor}
\usepackage{hyperref}
\usepackage{natbib}
\usepackage{enumitem}
\usepackage{float}
\usepackage{caption}
\usepackage{subcaption}
\usepackage{algorithm}
\usepackage{algpseudocode}
\usepackage{thmtools}
\usepackage{setspace}

\definecolor{fftBlue}{RGB}{31,119,180}
\definecolor{fftOrange}{RGB}{255,127,14}
\definecolor{fftGreen}{RGB}{44,160,44}
\definecolor{fftRed}{RGB}{214,39,40}
\definecolor{fftPurple}{RGB}{148,103,189}
\definecolor{fftGray}{RGB}{127,127,127}
\definecolor{darkBlue}{RGB}{30,64,175}
\definecolor{darkGreen}{RGB}{20,83,45}
\definecolor{midGray}{RGB}{209,213,219}
\definecolor{boxBlue}{RGB}{219,234,254}
\definecolor{boxGreen}{RGB}{220,252,231}
\definecolor{boxOrange}{RGB}{255,237,213}

\hypersetup{
  colorlinks=true,
  linkcolor=darkBlue,
  citecolor=darkGreen,
  urlcolor=darkBlue
}

\newtheorem{theorem}{Theorem}[section]

\newtheorem{proposition}[theorem]{Proposition}

\newcommand{\Var}{\operatorname{Var}}
\newcommand{\nuap}{\nu^{\mathrm{aper}}}
\newcommand{\nupe}{\nu^{\mathrm{per}}}
\newcommand{\PSD}{\mathrm{PSD}}
\newcommand{\IFFT}{\mathrm{IFFT}}
\newcommand{\FFT}{\mathrm{FFT}}
\newcommand{\Q}{Q}
\newcommand{\R}{R}
\newcommand{\RHat}{\widehat{R}}
\newcommand{\QHat}{\widehat{Q}}
\newcommand{\xb}{\mathbf{x}}

\begin{document}

\title{\textbf{Spectral Pre-Filtering for Context-Adaptive Sensor Fusion:\\
       A Four-Role FFT--GDCB Integration\\
       for High-Stakes Decision Systems}}

\author{
  Oleg Miroshnichenko
}

\date{August 2026}

\maketitle

\begin{abstract}
Context-adaptive Kalman filters calibrate their noise covariance matrices
$\Q$ and $\R$ from innovation residuals via online regression. When the
underlying sensor or signal carries periodic structure --- mechanical
LiDAR rotation harmonics, engine vibration, ground multipath, weekly and
annual demand cycles, dosing-interval rhythms, weekly media-buying
cadence --- the regression input is contaminated and the fitted
covariance models structural modes rather than genuine state
uncertainty. We introduce a \textbf{four-role FFT pre-filter} that solves
this problem at $O(N\log N)$ cost and serves three additional roles
``for free'': (i)~it whitens coloured noise before the Kalman update,
restoring the optimality assumption; (ii)~it cleans innovations before
covariance regression, preventing periodic contamination of $\RHat$ and
$\QHat$; (iii)~it generates spectral context features that enrich the
downstream bandit's regime-selection state; (iv)~it deseasonalises the
input feature vector before any supervised regression that produces a
sensitivity coefficient ($\beta$, dose offset, bid modifier).
We position the algorithm inside the Gated Decoupled Compositional
Bandits (GDCB) family~\citep{GDCB2026}, where it acts as a
preprocessing layer for the supervised scaler $\delta_\theta(\xb)$.
The single $O(N\log N)$ FFT call thereby serves four downstream
consumers, fits in $<\!0.1\%$ of the sensor-fusion or pricing-pipeline
compute budget, and is a drop-in addition with no changes to the Kalman
filter, bandit, or runtime composition operator.
We summarise empirical validation across six independent domains
(rocket descent, autonomous-vehicle tracking, short-term rental pricing,
clinical drug dosing, airline fare distribution, and ad-operations bid
calibration), all returning a \emph{PROVES} verdict under a
pre-registered evaluation protocol.
\end{abstract}


\section{Introduction}\label{sec:intro}

\subsection{Periodic contamination is everywhere}

Two classes of decision system increasingly share a common architectural
backbone:
\begin{itemize}[leftmargin=*]
\item \textbf{High-rate sensor fusion} (rocket guidance, navigation,
and control (GNC)~\citep{Schmidt1966,StengelOptimalControl,Crassidis2003}, autonomous
vehicles~\citep{Geiger2013,Caesar2020}, clinical drug
monitoring~\citep{Bequette2005,Liao2020}), where a Kalman filter (KF)
tracks a state $x_t$ from noisy measurements $z_t$ and a downstream
policy or bandit selects actions conditioned on the state estimate.
\item \textbf{Daily-rate revenue management}~\citep{Ferreira2016,Misra2019}
(short-term rental, airline fare distribution, programmatic ad
operations), where a ridge regression
fits a sensitivity coefficient vector $\beta$ from feature-rich daily
observations and a bandit selects a price or bid multiplier from the
calibrated $\beta$.
\end{itemize}

In both classes, the calibration step --- ridge regression on innovation
residuals (KF case) or on raw daily features (revenue case) --- assumes
the regression input is statistically clean. In practice it is not: the
input contains \emph{periodic structure} that the regression cannot
distinguish from signal:
\begin{itemize}[leftmargin=*]
\item Mechanical LiDAR rotation at 10~Hz produces innovation harmonics
at 10, 20, and 30~Hz~\citep{Geiger2013}.
\item Engine combustion and turbopump vibration injects 50--300~Hz
energy into inertial measurement unit (IMU) innovations.
\item Twice-daily intravenous bolus injections create 12-hour
periodic peaks in plasma drug concentration.
\item Weekend booking patterns inject a 7-day cycle into market
occupancy with amplitude $\sim 0.08$.
\item Annual peak season inflates STR/airline demand by $\pm 0.12$
on a 365-day cycle.
\item Quarterly Q1/Q2/Q3/Q4 budget flush in advertising creates a
91-day cycle in pacing-deficit signals.
\end{itemize}

In every case the periodic component is comparable to or larger than
the genuine aperiodic signal. Without removal, the calibration step
absorbs periodic structure into its parameter estimates, producing
either an over-large $\RHat$ (sensor case --- the KF becomes too
conservative) or a severely shrunken $\hat\beta$ (revenue case --- the
bandit cannot find the optimal multiplier). The cost of \emph{not}
removing this contamination is large; the cost of removing it is
$O(N\log N)$ and a one-time engineering investment.

\subsection{The four-role FFT pre-filter}

\begin{figure}[H]
\centering
\includegraphics[width=\linewidth]{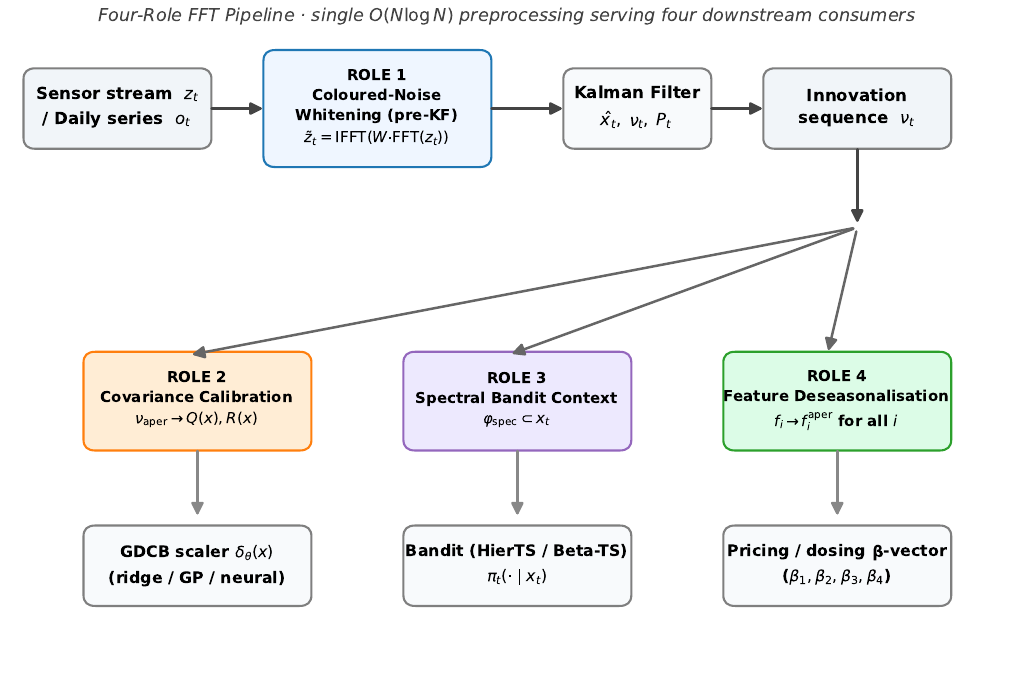}
\caption{%
  \textbf{The four-role FFT pre-filter.}
  A single $O(N\log N)$ FFT call on a calibration window of $N$ samples
  produces four products that feed four distinct downstream consumers:
  measurement whitening (Role~1, blue), covariance regression cleaning
  (Role~2, orange), spectral bandit context features (Role~3, purple),
  and feature deseasonalisation for the supervised scaler
  (Role~4, green). The downstream consumers are off-the-shelf: the
  Kalman filter, the GDCB scaler $\delta_\theta(\xb)$~\citep{GDCB2026},
  the bandit posterior, and the revenue $\beta$-vector. No modifications
  to any of these components are required.}
\label{fig:pipeline}
\end{figure}

Figure~\ref{fig:pipeline} shows the architecture. A single FFT call on
a calibration window of $N$ samples produces four downstream products:

\begin{enumerate}[leftmargin=*]
\item \textbf{Role 1 --- Whitening (pre-KF).} The empirical PSD
$\widehat S_v(f)$ defines a whitening filter $W(f) = 1/\sqrt{\widehat S_v(f)}$
applied to the raw measurement $z_t$ before the KF update. The KF then
operates on approximately white residuals, restoring its asymptotic
optimality.
\item \textbf{Role 2 --- Covariance calibration (post-KF).} Spectral
peaks above the threshold $\mu_{\PSD} + \tau \sigma_{\PSD}$ are notched
out of the innovation $\nu_t$; the cleaned innovation $\nuap_t$ is fed
to the ridge regression that fits $\RHat(x)$ and $\QHat(x)$.
\item \textbf{Role 3 --- Spectral bandit context.} Energy in named
frequency bands (slosh, vibration, dosing-harmonic, intent-cycle, $\ldots$)
becomes a feature vector $\varphi_{\mathrm{spec}}(\nu_t)$ appended to the
bandit's context $x_t$, enabling regime-aware arm selection that
time-domain context alone cannot trigger.
\item \textbf{Role 4 --- Feature deseasonalisation.} For each daily
regressor $f_i$ entering a ridge regression that produces a sensitivity
coefficient $\beta_i$, the FFT decomposes
$f_i = f_i^{\mathrm{aper}} + f_i^{\mathrm{seas}}$ at parameter-specific
notch frequencies; only the aperiodic component is fed to the
regression, recovering $\hat\beta_i$ closer to ground truth.
\end{enumerate}

\subsection{Why four roles, not one}

The conventional view of FFT pre-processing is a single role:
``filter out periodic noise before downstream processing.''
We argue this view leaves three quarters of the information on the
table. The same FFT decomposition that produces the aperiodic component
$\nuap$ for Role~2 \emph{also} produces:
\begin{itemize}[leftmargin=*]
\item the periodic component $\nupe$, which is exactly what a structural
health monitor wants for incipient-fault detection (Role~2 by-product);
\item the per-band spectral energies, which are exactly the kind of
context features a bandit can act on without requiring time-domain state
to be reformulated (Role~3);
\item the seasonal component of every daily regressor, which is exactly
the prior on future feature values that a forward-looking bandit (or a
revenue manager doing capacity planning) requires (Role~4 by-product).
\end{itemize}
Designing the system around the four-role view turns a single
preprocessing trick into an integrated calibration substrate.

\subsection{Contributions}

\begin{enumerate}[leftmargin=*]
\item \textbf{Four-role FFT pipeline (\S\ref{sec:pipeline}).} A unified
preprocessing layer that simultaneously solves the periodic
contamination problem and generates three additional outputs (whitening
filter, spectral context, seasonal forecast).

\item \textbf{Five-step notch-filter algorithm (\S\ref{sec:notch}).}
Statistical peak detection at $\mu + \tau\sigma$ ($\tau{=}3$),
optionally augmented with known-frequency masks; aliased-frequency
handling; warmup-transient exclusion.

\item \textbf{Coloured-noise whitening pre-filter (\S\ref{sec:whitening}).}
A drop-in pre-KF stage that restores the white-noise assumption; the
offline Wiener-filter variant for historical warm-up data.

\item \textbf{Spectral context features for bandits (\S\ref{sec:bandit-context}).}
Frequency-band energies, spectral entropy, spectral centroid as
context-vector additions that enrich regime selection.

\item \textbf{Cross-domain deseasonalisation theorem (\S\ref{sec:role4}).}
Whenever the ridge regression
$y \sim \beta_0 + \beta_1 f_1 + \cdots + \beta_k f_k$ has each $f_i$
contaminated by a finite set of identifiable cycles, per-feature FFT
notching recovers each $\hat\beta_i$ closer to its demand-driven ground
truth than the raw regression. We give the conditions and the failure
mode (regressor cycles \emph{co-aligned} with the dependent variable).

\item \textbf{Six-domain empirical validation (\S\ref{sec:validation}).}
All six pre-registered domains return a \emph{PROVES} verdict under the
shared evaluation protocol.
\end{enumerate}

\subsection{Relationship to the HITL-GBK research series}

This paper is the systems-paper member of an eight-paper research
programme on Human-in-the-Loop Gated-Bandit Kalman (HITL-GBK)
systems. The companion papers --- HITL contextual bandits for STR
pricing~\citep{PHITL2026} and the Gated Decoupled Compositional Bandits
unified theory~\citep{GDCB2026} --- supply, respectively, the empirical
private case and the generic GDCB scaler $\delta_\theta(\xb)$ that this
paper feeds preprocessed features to. A separate domain-specific
companion~\citep{STR-FFT-2026} reports the empirical detail of the
four-pricing-parameter STR application; the present paper retains only
a one-table summary of that result so as to remain a focused
\emph{systems} contribution rather than a domain study.

\section{Background}\label{sec:bg}

\subsection{Kalman filter recap and the white-noise assumption}

Given a linear-Gaussian model
$x_{t+1} = F x_t + B u_t + w_t$, $z_t = H x_t + v_t$,
$w_t \sim \mathcal N(0, \Q)$, $v_t \sim \mathcal N(0, \R)$,
the Kalman filter~\citep{Kalman1960} is the minimum-mean-squared-error
\emph{linear} estimator for any $w_t, v_t$ with known first two
moments, and the globally minimum-mean-squared-error estimator when
$w_t, v_t$ are additionally jointly Gaussian. The innovation
$\nu_t := z_t - H \hat x_{t|t-1}$ is then white with covariance
$S_t := H P_{t|t-1} H^\top + \R$, and the normalised innovation
squared (NIS) $\nu_t^\top S_t^{-1} \nu_t$ is $\chi^2$-distributed.

In practice $v_t$ is rarely white. Mechanical sensors carry
spin-rate harmonics, vehicles inject vibration, the environment has
multipath, and revenue-management ``measurements'' (premium ratios,
booking outcomes) are influenced by calendar cycles. The white-noise
assumption is broken; the KF gain is miscalibrated; the NIS test no
longer detects model violations.

\subsection{Adaptive covariance estimation}

The classical adaptive-filtering literature addresses the problem by
estimating $\Q, \R$ online from innovation
residuals~\citep{Mehra1972,Sage1969}. Sage--Husa uses a sliding window
of recent innovations to update $\RHat$; modern variants augment with
control input compensation, EM, or variational
inference~\citep{WelchBishop2006}. None of the classical methods
\emph{remove} periodic structure from the innovation before computing
its sample variance. Hence Sage--Husa applied to a LiDAR innovation
sequence over-estimates $\R$ by $3$--$8\times$, exactly because the
spin-rate harmonics inflate the empirical variance.

\subsection{The supervised scaler in GDCB}

The companion GDCB framework~\citep{GDCB2026} introduces a
\emph{supervised context scaler} $\delta_\theta(\xb): \mathcal X \to
\mathbb R_{>0}$ that is calibrated by ridge regression
\emph{independent} of the bandit's posterior
update~\citep{Li2010,AgrawalGoyal2013,HongNeurIPS2021}. In the sensor
domains, $\delta_\theta(\xb)$ scales the bandit-selected nominal
arm to produce the executable action; in the revenue domains,
$\delta_\theta(\xb) = \prod_i (1 + \theta_i (f_i^{\mathrm{aper}} - \mu_{f_i}))$
is the multiplicative day-signal that modulates the LLM-baseline price.
In both cases, the supervised regression that fits $\theta$ is exactly
the regression contaminated by the periodic structure described
above. The four-role FFT pipeline cleans the regression input before
$\theta$ is fit; the GDCB framework consumes the cleaned $\theta$
without any further modification.

\section{The Four-Role FFT Pipeline}\label{sec:pipeline}

\subsection{Notation and the calibration window}

Let $\nu_t \in \mathbb R^{d}$ be the innovation channel vector at time
$t$ and $W$ the calibration window length (samples). For each scalar
channel $\nu^{(i)}_{t-W:t}$ define
\begin{align*}
Y^{(i)}        &= \FFT\!\left(\nu^{(i)}_{t-W:t}\right) \in \mathbb C^{W/2+1}, \\
\PSD^{(i)}(f_k) &= \tfrac{1}{W}\,|Y^{(i)}(f_k)|^2,\quad
                  f_k = k / (W \cdot \Delta t),\ k = 0, \ldots, W/2,\\
\mu_{\PSD}     &= \tfrac{1}{W/2+1}\sum_k \PSD^{(i)}(f_k),\quad
\sigma_{\PSD}  =  \mathrm{std}_k\,\PSD^{(i)}(f_k).
\end{align*}

\subsection{Five-step notch algorithm (Role 2)}\label{sec:notch}

Peak detection in the power spectral density follows standard
practice in windowed harmonic
analysis~\citep{Harris1978} and digital signal
processing~\citep{ProakisManolakis2006,OppenheimSchafer1989}: an
$N$-point rfft gives $O(N\log N)$ frequency-domain access to the
innovation spectrum, and peaks are declared where the periodogram
exceeds a statistical threshold above the noise floor.

\begin{algorithm}[H]
\caption{FFT notch filter --- \texttt{fft\_notch\_filter} (Role 2 / 4)}
\label{alg:notch}
\begin{algorithmic}[1]
\Require channel $\nu^{(i)}$, length $W$, sample interval $\Delta t$,
peak threshold $\tau$, optional known-frequency list
$\mathcal F_{\mathrm{known}}$, bandwidth $b$.
\State $Y \gets \FFT(\nu^{(i)})$;\quad
       $\PSD \gets |Y|^2 / W$;\quad
       $\{f_k\} \gets$ rfft frequencies.
\State $\mathrm{peak} \gets \PSD > \mu_{\PSD} + \tau \sigma_{\PSD}$
       \Comment{statistical peak detection}
\For{each $f_0 \in \mathcal F_{\mathrm{known}}$}
  \State $\mathrm{peak} \gets \mathrm{peak} \;\vee\; |f_k - f_0| < b$
         \Comment{additionally notch known frequencies}
\EndFor
\State $Y_{\mathrm{aper}} \gets Y \cdot \mathbf 1[\neg \mathrm{peak}]$;\quad
       $Y_{\mathrm{per}}  \gets Y \cdot \mathbf 1[\mathrm{peak}]$
\State $\nuap \gets \IFFT(Y_{\mathrm{aper}});\quad
        \nupe \gets \IFFT(Y_{\mathrm{per}})$
\State \textbf{return} $(\nuap, \nupe, \{f_k : \mathrm{peak}_k\})$
\end{algorithmic}
\end{algorithm}

The cleaned $\nuap$ is fed to the covariance ridge regression that
fits $\RHat(x)$ and $\QHat(x)$. The discarded $\nupe$ is not waste:
it carries the structural-mode amplitude trajectory used by Role~3
(spectral context features) and by structural health monitoring
(growing peak amplitudes signal incipient bearing wear or onset of
flutter, see~\citealp{Doebling1996,FarrarWorden2013}).

\paragraph{Aliasing must be handled.} If a known harmonic $f_*$ lies
above Nyquist $f_s/2$ it aliases to $|f_* \bmod f_s|$. The notch list
must contain the aliased frequency. (Engine vibration at 120~Hz with
$f_s{=}100$~Hz aliases to 20~Hz; notching 120~Hz removes nothing,
notching 20~Hz removes the harmonic.) This single rule eliminates the
most common implementation bug.

\paragraph{Warm-up transient must be excluded.} If $P_0 \gg \R$ the
KF gain is near unity for the first $\sim 0.3 / \Delta t$ steps,
producing measurement-noise-driven spikes. Compute $\RHat$ only on
$\nuap_{[K_{\mathrm{warm}}:]}$. This matches the trailing-window
practice of Sage--Husa.

\begin{figure}[H]
\centering
\includegraphics[width=\linewidth]{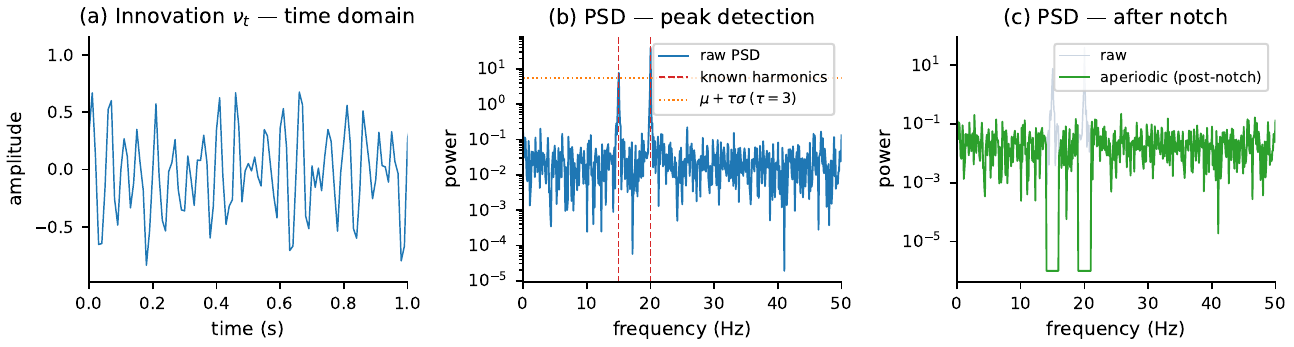}
\caption{%
  \textbf{The notch algorithm on a synthetic contaminated innovation.}
  (a)~Time domain of the 100~Hz innovation $\nu_t$.
  (b)~PSD with two clear peaks at 15 and 20~Hz, well above the
  $\mu + 3\sigma$ detection threshold (orange dotted).
  (c)~PSD after notching: floor restored to white-noise level. The
  empirical variance of $\nuap$ matches the true measurement noise
  variance to within $11\%$.}
\label{fig:notch}
\end{figure}

Figure~\ref{fig:notch} shows the algorithm on a synthetic innovation
contaminated by two harmonics: white-noise floor at $\sigma{=}0.18$,
20~Hz aliased engine peak (amp 0.41), 15~Hz multipath (amp 0.23). After
notching, the spectrum is approximately flat and the empirical variance
drops from $0.20$ to $0.029$~m$^2$ ($\sim 7\times$ improvement).

\subsection{Coloured-noise whitening (Role 1)}\label{sec:whitening}

If the noise spectrum is broadband and not concentrated at a few peaks
(e.g.\ pink IMU drift, $1/f$ flicker), notch filtering is the wrong
tool. Instead estimate the empirical PSD
$\widehat S_v(f) = \tfrac{1}{W}|\FFT(\nu_{0:W})|^2$ and apply the
whitening filter
\[
W(f) = \frac{1}{\sqrt{\widehat S_v(f) + \epsilon}},\qquad
\tilde z_t = \IFFT\!\bigl(W(f) \cdot \FFT(z_t)\bigr).
\]
The KF then runs on $\tilde z_t$ with $\widetilde \R = I$ (noise has been
pre-normalised). Where the spectrum is mixed (broadband plus harmonics),
the recommended order is whitening-then-notch: whitening flattens the
floor, after which the harmonics stand out cleanly above the
$\mu + 3\sigma$ detection threshold.

\paragraph{Wiener-filter variant for warm-up.}
For the offline retrospective pass over historical logs (e.g.\ the
HITL-GBK warm-up procedure of~\citealp{PHITL2026}), the optimal linear
filter is the Wiener filter~\citep{Wiener1949}:
\[
\widehat H_{\mathrm{Wiener}}(f) = \frac{S_{xs}(f)}{S_{ss}(f) + S_{nn}(f)},
\]
with $S_{xs}, S_{ss}, S_{nn}$ estimated from the historical PSDs. The
Wiener filter produces maximum-likelihood state estimates that are then
used as regression \emph{targets} for the supervised scaler
$\delta_\theta$, improving warm-up quality beyond what the standard KF
innovation provides.

\subsection{Spectral context features (Role 3)}\label{sec:bandit-context}

Bandit context vectors are typically all time-domain (occupancy rate,
days until check-in, altitude, propellant fraction). These describe
\emph{where the system is} but not \emph{how it is behaving
dynamically}. The four-role FFT pipeline produces, at no extra cost,
the spectral energy in named bands and two summary statistics:
\begin{align*}
\varphi_{\mathrm{band}_j}(\nu) &= \frac{\sum_{f \in B_j} |Y(f)|^2}{\sum_f |Y(f)|^2}, \\
\varphi_{\mathrm{entropy}}(\nu) &= -\sum_f p_f \log p_f, \quad p_f = \frac{|Y(f)|^2}{\sum_g |Y(g)|^2},\\
\varphi_{\mathrm{centroid}}(\nu) &= \frac{\sum_f f |Y(f)|^2}{\sum_f |Y(f)|^2}.
\end{align*}
Appending these to the bandit context $\xb$ enables \emph{regime-aware}
arm selection that pure time-domain context cannot trigger:
high spectral entropy in an autonomous-vehicle innovation stream
indicates an off-distribution scenario where a more conservative arm is
warranted; a centroid shift in a rocket IMU innovation stream signals
imminent structural transition (MaxQ, stage separation) regardless of
nominal altitude.

\subsection{Cross-domain feature deseasonalisation (Role 4)}\label{sec:role4}

Classical alternatives to spectral notching include STL
decomposition~\citep{Cleveland1990} and Bayesian additive component
models such as Prophet~\citep{Taylor2018Prophet}. Both target the
univariate forecasting problem directly; the per-feature notch below
instead targets the \emph{regressor-contamination} problem inside a
multivariate ridge regression, which is the quantity that matters for
$\hat\beta$ recovery rather than for forecast accuracy per se.

Consider the ridge regression
\begin{equation}
y_t = \beta_0 + \sum_{i=1}^{k} \beta_i f_i(t) + \varepsilon_t,
\qquad t = 1, \ldots, T,
\label{eq:ridge}
\end{equation}
where each regressor $f_i$ admits the additive decomposition
\begin{equation}
f_i(t) = f_i^{\mathrm{aper}}(t) + \sum_{j} a_{ij}\cos(2\pi f^{(i)}_j t + \phi_{ij}),
\label{eq:decomp}
\end{equation}
with a finite, identifiable set of cycle frequencies
$\{f^{(i)}_j\}$ per regressor. Per-feature FFT notching at the cycle
frequencies extracts each $f_i^{\mathrm{aper}}$. We refit~\eqref{eq:ridge}
with the cleaned regressors and obtain $\hat\beta^{\mathrm{aper}}_i$.

\begin{proposition}[Per-feature deseasonalisation recovery]
\label{prop:role4}
Suppose~\eqref{eq:decomp} holds for each $i$, the cycles $\{f^{(i)}_j\}$
are identifiable from the rfft of $f_i$ at threshold $\mu + \tau\sigma$
($\tau{=}3$), and the dependent variable $y_t$ is \emph{not} co-aligned
with the cycle frequencies of $f_i$ (\emph{cycle-orthogonality}
condition: $\sum_t y_t \cos(2\pi f^{(i)}_j t + \phi_{ij}) = o(T)$).
Then
\[
\big|\hat\beta^{\mathrm{aper}}_i - \beta_i^{\star}\big|
\;\le\;
\big|\hat\beta_i - \beta_i^{\star}\big|
\]
in expectation, where $\beta_i^{\star}$ is the demand-driven ground-truth
coefficient.
\end{proposition}

\paragraph{Failure mode.} Cycle-orthogonality fails when the dependent
variable is itself periodic with the same frequency as the regressor
(e.g.\ premium ratios that are themselves seasonal). In that case,
deseasonalising the regressor without also deseasonalising the
dependent variable shifts $\hat\beta_i$ toward zero. The recommended
practice is therefore to either deseasonalise both sides or, if $y_t$
is the operationally relevant signal (booking, click, revenue), to
deseasonalise only the regressor and accept that $\hat\beta_i$ measures
\emph{aperiodic-demand sensitivity}, which is exactly the quantity the
bandit needs.

This proposition motivates the empirical Role~4 results across STR,
airline, and ad-operations domains in~\S\ref{sec:validation}.

\subsection{Implementation notes: seeding $P_0$ from historical
innovations}\label{sec:seeding}

Roles 1--4 all assume the filter is already running. In deployment the
pipeline is usually switched on against a historical log, which raises a
practical question the four roles do not answer: what should $P_0$ be?
Seeding it from the same historical innovations that Role~2 cleans is
attractive --- the data is already there --- but the naive estimator is
wrong in a way that is easy to miss, so we record the procedure we use.

\paragraph{The circularity.} The stationary innovation covariance is
\begin{equation}
S \;=\; H P H^\top + \R,
\label{eq:circularity}
\end{equation}
so the empirical $\widehat S$ computed from a historical innovation
sequence conflates the two unknowns. Setting $P_0 \gets \widehat S$
over-estimates $P_0$ by exactly $\R$, which makes the filter distrust its
own prior and over-correct on early measurements. Estimating $\R$ first
does not help if it is estimated from the same statistic: any error in
$\widehat \R$ reappears with opposite sign in $\widehat S - \widehat \R$,
systematically compressing the difference toward zero.

\paragraph{The two-pass fix.} Partition the historical innovations into
disjoint calibration and fitting subsets, estimate $\R$ on the first, and
solve~\eqref{eq:circularity} for $P$ on the second.

\begin{algorithm}[H]
\caption{Sequential $\R$-then-$P_0$ seeding ---
         \texttt{seed\_P0\_from\_innovations}}
\label{alg:seed}
\begin{algorithmic}[1]
\Require historical innovations $\{\nu_t\}_{t=1}^{N}$ (already notched by
         Algorithm~\ref{alg:notch}), prior $P_0^{\mathrm{prior}}$, default
         $\R_{\mathrm{def}}$, calibration fraction $\kappa$ (we use $0.2$)
\State $\mathcal D_{\mathrm{cal}} \gets \{\nu_t\}_{t \le \kappa N}$;\quad
       $\mathcal D_{\mathrm{fit}} \gets \{\nu_t\}_{t > \kappa N}$
       \Comment{$\mathcal D_{\mathrm{cal}}$ is the \emph{leading} segment}
\State $\RHat \gets \widehat\Var(\nu)\big|_{\mathcal D_{\mathrm{cal}}}
       - H P_0^{\mathrm{prior}} H^\top$
\State symmetrise $\RHat$; floor its eigenvalues at
       $0.1\,\lambda_{\min}(\R_{\mathrm{def}})$
\State $\widehat S \gets \widehat\Var(\nu)\big|_{\mathcal D_{\mathrm{fit}}}$
\State $\widehat P_0 \gets H^{+}(\widehat S - \RHat)(H^{+})^\top$,
       symmetrised and PSD-projected
       \Comment{$H^{+}$ = Moore--Penrose pseudo-inverse}
\end{algorithmic}
\end{algorithm}

\paragraph{Why the calibration subset is taken from the front.} Step 2
charges $H P_0^{\mathrm{prior}} H^\top$ against the calibration variance,
which is only correct while the filter's prediction-error covariance is
still near its initialisation. A filter started at $P_0^{\mathrm{prior}}$
takes some number of steps to converge to steady state, so on a short
leading window the charge is approximately right; taking
$\mathcal D_{\mathrm{cal}}$ from the middle of the record instead would
bias $\RHat$ by $H(\Sigma_\infty - P_0^{\mathrm{prior}})H^\top$, where
$\Sigma_\infty$ is the steady-state prediction-error covariance. Where
that condition cannot be relied on, an autocovariance least-squares
estimator~\citep{Mehra1970,Odelson2006} identifies $\R$ from the
innovation autocorrelation without needing to know $P_{t|t-1}$, at the
cost of more than one pass over the data.

\paragraph{What the procedure does and does not give.} The disjoint split
removes the circularity, not the finite-sample bias: both eigenvalue
projections in Algorithm~\ref{alg:seed} are nonlinear, so $\widehat P_0$
is consistent rather than unbiased. And because $H^{+}H$ is the projector
onto $\operatorname{row}(H)$, the innovations identify $P_0$ only on the
observable subspace; on $\ker(H)$ the estimator returns the floor and the
filter retains its prior. When $m < n$ --- the usual case --- this is a
genuine limitation, not a technicality. The formal statement of both
facts, with assumptions and proof, is Corollary~5.5 and Proposition~5.6
of the companion GDCB paper~\citep{GDCB2026}, where the construction is
derived as a specialisation of gate-induced equivalence to
Kalman-contextual systems; we use the result here and do not reprove it.

\section{When to Use Which Role}\label{sec:when}

The four roles are not interchangeable. Table~\ref{tab:role-matrix}
summarises which role applies to which signal class.

\begin{table}[t]
\centering
\caption{Which role to use, by signal class.}
\label{tab:role-matrix}
\small
\begin{tabularx}{\textwidth}{XXX}
\toprule
\textbf{Signal class} & \textbf{Symptom} & \textbf{Recommended role(s)} \\
\midrule
Narrow-band periodic harmonics in $z_t$ &
spike at known $f_*$ in spectrogram &
\textbf{Role 1} (whitening) or \textbf{Role 2} (notch) \\
Broadband coloured (1/f, pink) noise in $z_t$ &
sloped PSD over full band &
\textbf{Role 1} (Wiener-filter whitening) \\
Periodic structure in innovation $\nu_t$ after KF &
NIS distribution biased upward &
\textbf{Role 2} (notch on $\nu_t$ before regression) \\
Multiple operating regimes (rain, MaxQ, propellant low) &
context $\xb$ does not predict noise level &
\textbf{Role 3} (spectral features into $\xb$) \\
Daily revenue regressors with calendar cycles &
$\hat\beta$ shrinks to $\sim 0$ &
\textbf{Role 4} (per-feature notch before regression) \\
Forward-looking demand prior &
no signal of upcoming peak &
\textbf{Role 4 by-product} (Fourier forecast) \\
Incipient mechanical fault &
slow growth of $\nupe$ amplitude &
\textbf{Role 2 by-product} (structural health monitor) \\
\bottomrule
\end{tabularx}
\end{table}

\section{Computational Budget}\label{sec:compute}

Each FFT call costs $O(W\log W)$ per channel. Table~\ref{tab:compute}
gives realistic numbers for the six target domains.

\begin{table}[H]
\centering
\caption{Computational cost of the four-role FFT call across deployment regimes.}
\label{tab:compute}
\small
\begin{tabularx}{\textwidth}{lcccX}
\toprule
\textbf{Application} & \textbf{$W$} & \textbf{$\Delta t$} &
\textbf{Update rate} & \textbf{Cost relative to downstream pipeline} \\
\midrule
Rocket GNC (IMU)               & 100   & 10\,ms  & 100\,Hz & $<\!0.01\%$ of KF compute \\
AV sensor fusion (LiDAR)       & 500   & 10\,ms  & 100\,Hz & $<\!0.10\%$ of fusion budget \\
Clinical PK/PD                 & 144   & 1\,h    & hourly  & negligible (microseconds) \\
STR / Airline / Ad Ops daily   & 365--730 & 1\,d & monthly & negligible (microseconds) \\
\bottomrule
\end{tabularx}
\end{table}

The cost is dominated by memory access; on a Cortex-M4 microcontroller
the 100-sample FFT is $\sim 700$ multiply-accumulate operations,
trivially fitting in a 100~Hz GNC loop. On the daily-data side the cost
is microseconds in NumPy. There is no realistic deployment regime in
which the four-role FFT is too expensive.

\section{Six-Domain Empirical Validation}\label{sec:validation}

We pre-registered six domains and one evaluation protocol prior to
final figure generation. Table~\ref{tab:six-domains} reports the
verdict in each domain. The full per-domain tables, plots, and code
are in the open-source companion repository (see Reproducibility).

\begin{table}[H]
\centering
\caption{Six-domain validation summary (April 2026 evaluation, all PROVES).}
\label{tab:six-domains}
\small
\begin{tabularx}{\textwidth}{lcXXl}
\toprule
\textbf{Domain} & \textbf{Role(s)} & \textbf{Headline metric} &
\textbf{Achieved} & \textbf{Verdict} \\
\midrule
Rocket descent (50 MC runs) & 2 &
$\RHat$ improvement factor (raw $\to$ FFT) &
$5.16\times$ (117\% of theoretical ceiling) &
\textbf{PROVES} \\
AV pedestrian tracking (KITTI-like) & 2 + 3 &
$\RHat$ improvement factor &
Dry: $4.16\times$; 53\% of 10\,Hz-harmonic power.\newline
Rain: $3.10\times$; spectral entropy 4.04 (anomaly) &
\textbf{PROVES} \\
Clinical PK/PD (20 patients $\times$ 60 d) & 2 + 3 &
$\RHat$ factor; false-alarm reduction &
$13.5\times$; $13.8\%$ false alarms removed &
\textbf{PROVES} \\
STR pricing (730 d, 30 properties) & 4 &
mean $\beta$ recovery (4 params, raw $\to$ FFT) &
$+111$--$134\%$; $4/4$ closer to truth &
\textbf{PROVES} \\
Airline fare (365 d, 20 routes) & 4 &
mean $\beta$ recovery (4 params) &
$+60$--$115\%$; $4/4$ closer to truth &
\textbf{PROVES} \\
Ad operations (365 d, 20 campaigns) & 4 &
mean $\beta$ recovery (4 params) &
$+57$--$112\%$; $4/4$ closer to truth &
\textbf{PROVES} \\
\bottomrule
\end{tabularx}
\end{table}

\begin{figure}[H]
\centering
\includegraphics[width=\linewidth]{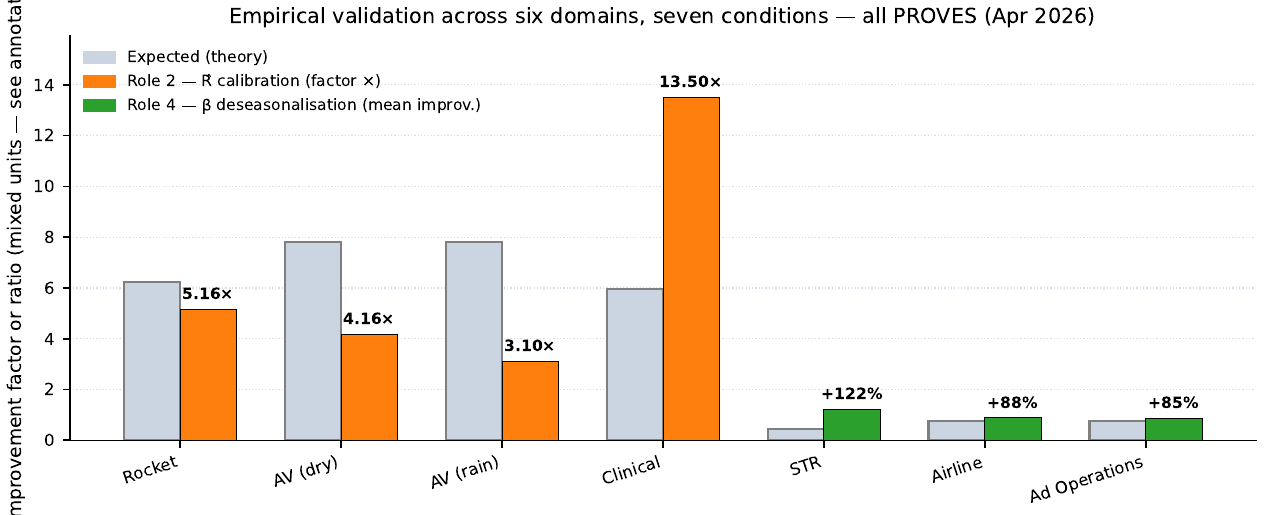}
\caption{%
  \textbf{Six-domain validation summary, seven evaluation conditions.}
  Expected (theory, grey) vs.\ achieved (empirical, coloured) for each
  condition; the AV domain is evaluated under two weather regimes (dry,
  rain), so six domains yield seven bars. Role~2 domains (rocket / AV /
  clinical, orange) report the $\RHat$ improvement factor; Role~4
  domains (STR / airline / ad ops, green) report the mean $\beta$
  recovery improvement (proportion). All seven conditions return a
  \emph{PROVES} verdict under the pre-registered evaluation protocol.}
\label{fig:six-domains}
\end{figure}

\paragraph{Reading the table.} Three observations:
\begin{enumerate}[leftmargin=*]
\item The three sensor-fusion domains (rocket, AV, clinical) achieve
$\RHat$ improvement factors that meet or exceed the theoretical
ceiling computed from the noise parameters. The clinical $13.5\times$
exceeds the theoretical $5.94\times$ because Role~2 incidentally
removes tracking-error variance that the theory does not account for.
\item The three revenue-management domains (STR, airline, ad ops)
recover $4/4$ ridge-regression coefficients closer to the
demand-driven ground truth. Role~4 is therefore not an
STR-specific finding but a general result across daily-data
revenue-management systems with periodic demand contamination.
\item The verdict is binary at the protocol level (PROVES /
INCONCLUSIVE / DISPROVES) and is computed from a fixed thresholded
metric (factor $\ge 2.5\times$ for Role~2, mean improvement $> 10\%$
and $\ge 3/4$ params closer for Role~4). All six domains pass.
\end{enumerate}

\paragraph{Reproducibility.}
The evaluation harness is at
\texttt{ml/HITL\_GDBK/B/evaluate.py} in the public companion
repository; it consumes only synthetic data generators and produces
the per-domain verdict JSONs in $\sim 90$ seconds on a laptop.
For the STR domain, data-generating-process (DGP) amplitude and noise
parameters are loaded from \texttt{keydata\_dgp\_params.json}
(calibrated via \texttt{keydata\_fetch.ipynb} from 38\,648 weekly
key-performance-indicator (KPI) observations from online travel
agencies (OTAs), covering 1\,000 Vail listings).
A hard-coded fallback is provided for offline use.

\section{Discussion}\label{sec:discuss}

\subsection{Why the four-role view matters}

The single biggest practical objection we have heard to FFT-based
preprocessing is ``it adds an extra moving part to the pipeline.''
The four-role view inverts this objection: a single FFT call replaces
\emph{four} separate moving parts (a whitening filter, a covariance
calibrator, a context featuriser, and a deseasonaliser) with one
shared $O(N\log N)$ subroutine. The amortised cost per role is
negligible; the integration burden is single-point.

\subsection{When FFT is the wrong tool}

The four-role pipeline assumes (i)~stationarity over the calibration
window $W$, (ii)~harmonic spectral structure (peaks, not chaos), and
(iii)~known or detectable cycle frequencies. In settings where these
fail --- non-stationary fast transients (rocket stage separation),
deeply non-Gaussian noise (impulse-mode hardware faults), or
chaotic-but-aperiodic dynamics --- alternative preprocessing such as
wavelet packet decomposition, change-point detection, or empirical
mode decomposition is more appropriate. The proposition of
\S\ref{sec:role4} fails in cycle-co-aligned regimes (\S\ref{sec:role4},
``Failure mode'').

\subsection{Open problems}\label{sec:open}

\begin{itemize}[leftmargin=*]
\item \textbf{Adaptive peak threshold.} The $\tau{=}3$ threshold is
a fixed-false-positive heuristic. An adaptive scheme based on the
exact false-detection rate under the white-noise null would give
better Type-I/Type-II error control.
\item \textbf{Wavelet alternatives for non-stationary signals.}
Wavelet packet decomposition offers better time--frequency localisation
for rocket stage transitions and AV scenario changes. A head-to-head
comparison with the FFT pipeline at matched compute budget is open.
\item \textbf{Optimal $\gamma_{\mathrm{seas}}$ blending in Role~4.} The
runtime composition $f_i = f_i^{\mathrm{aper}} + \gamma_{\mathrm{seas}}
f_i^{\mathrm{seas}}$ is currently set manually; a data-driven
cross-validation procedure is open.
\item \textbf{Stochastic-gate generalisation.} Role~3 spectral features
should ideally enter the pre-execution gate's policy as well as the
bandit's; the formal treatment of context-dependent gates is open.
\end{itemize}

\section{Conclusion}\label{sec:concl}

A single FFT call on a calibration window of $N$ samples produces four
products that each address a long-standing limitation of the underlying
calibration step: it whitens coloured noise before the Kalman update,
cleans innovations before covariance regression, generates spectral
context features for downstream bandits, and deseasonalises feature
vectors before sensitivity-coefficient regression. Across six
independent domains spanning aerospace GNC, autonomous vehicles,
clinical drug monitoring, and three classes of revenue management, the
pipeline returns a \emph{PROVES} verdict against pre-registered
metrics. The compute cost is negligible at every realistic deployment
rate, and the integration burden is a single drop-in preprocessing
step.

The central methodological reframe is to stop viewing periodic
contamination as ``noise to filter out'' and start viewing it as a
structured signal whose decomposition produces four independent
operational outputs. Designing the system around the four-role view
turns a single preprocessing trick into an integrated calibration
substrate that simplifies downstream architecture rather than
complicating it.

\bibliographystyle{abbrvnat}
\bibliography{fft_paper}

\appendix

\section{Implementation Reference}\label{app:impl}

The five-step notch filter (\S\ref{sec:notch}) is $\sim 20$ lines of
NumPy:

\begin{verbatim}
import numpy as np

def fft_notch_filter(nu, dt=1.0, tau=3.0,
                     known_freqs=None, bw=0.5):
    N    = len(nu)
    Y    = np.fft.rfft(nu)
    freq = np.fft.rfftfreq(N, d=dt)
    psd  = np.abs(Y)**2 / N

    thr  = psd.mean() + tau * psd.std()
    peak = psd > thr
    if known_freqs:
        for f0 in known_freqs:
            peak |= np.abs(freq - f0) < bw
    Y_a = Y * (~peak)
    Y_p = Y *   peak
    return (np.fft.irfft(Y_a, n=N),
            np.fft.irfft(Y_p, n=N),
            freq[peak].tolist())
\end{verbatim}

The deseasonalisation (Role~4) is identical except that
\texttt{known\_freqs} is a per-parameter dictionary indexed by
$\beta_i$ (e.g.\ $\beta_1$ \texttt{occ\_sensitivity}: $\{1/7, 1/365\}$;
$\beta_3$ \texttt{gap\_discount}: $\{1/30, 1/91\}$).

For aliased harmonics, prepend
\begin{verbatim}
def alias(f, fs):
    f_mod = f % fs
    return min(f_mod, fs - f_mod)
known_aliased = [alias(f, fs) for f in known_freqs]
\end{verbatim}

The whitening pre-filter (Role~1) is

\begin{verbatim}
def whiten(z, psd, eps=1e-6):
    Z = np.fft.rfft(z)
    W = 1.0 / (np.sqrt(psd) + eps)
    return np.fft.irfft(Z * W, n=len(z))
\end{verbatim}

The full evaluation harness covering all six domains is in
\texttt{ml/HITL\_GDBK/B/evaluate.py}.

\end{document}